\PassOptionsToPackage{hyphens}{url}
\documentclass{modelica}
\usepackage{microtype}

\usepackage[table]{xcolor}

\hypersetup{
  pdftitle  = {RL for TCL Control using Modelica and Python},
  pdfauthor = {Oleh Lukianykhin, Tetiana Bogodorova},
  pdfsubject = {Asian Modelica Conference 2020},
  pdfkeywords = {Modelica, confercence, LaTeX, template},
  colorlinks,
  linkcolor=black,
  urlcolor=black,
  citecolor=black,
  pdfpagelayout = SinglePage,
  pdfcreator = pdflatex,
  pdfproducer = pdflatex}

\begin{document}
\thispagestyle{empty}

\title{Reinforcement Learning for Thermostatically Controlled Loads Control using Modelica and Python}
\author[1]{Oleh Lukianykhin}
\author[2]{Tetiana Bogodorova}
\affil[1]{The Machine Learning Lab, Ukrainian Catholic University, Ukraine, {\small\texttt{lukianykhin@ucu.edu.ua}}}
\affil[2]{Department of Electrical, Computer, and System Engineering, Rensselaer Polytechnic Institute, Troy, NY, USA {\small\texttt{bogodt2@rpi.edu}}}

\maketitle\thispagestyle{empty} 
\abstract{
The aim of the project is to investigate and assess opportunities for applying reinforcement learning (RL) for power system control. As a proof of concept (PoC), voltage control of thermostatically controlled loads (TCLs) for power consumption regulation was developed using Modelica-based pipeline. The Q-learning RL algorithm has been validated for deterministic and stochastic initialization of TCLs. The latter modelling is closer to real grid behaviour, which challenges the control development, considering the stochastic nature of load switching. In addition, the paper shows the influence of Q-learning parameters, including discretization of state-action space, on the controller performance.
\footnote{Experiment pipeline, procedure, full results, visualizations and analysis are available at \url{https://github.com/OlehLuk/rl-power-control}}}

\noindent\emph{Keywords: Modelica, Dymola, Open AI Gym, JModelica.org, OpenModelica, Python, Reinforcement Learning, Q-learning, Thermostatically Controlled Loads, Power System, Demand Response}

\section{Introduction}


Despite of the successful application in the past, classic methods and solutions in power systems are not capable to handle new challenges. In particular, when stability margins have decreased due to stochastic behaviour of renewable energy sources that increase presence in the power grid \cite{begovic2001impact}. In addition, the rise of IoT-related technologies contributed to appearance of distributed smart grid \cite{ipakchi2009grid}. These challenges require and allow for new solutions, one of those is an application of reinforcement learning algorithms for controller design.

The reinforcement learning learns from the interaction of a controller (agent) with a system (environment). Examples of successful applications for complex tasks in various domains are winning complex games \cite{alphago,alphastarblog}, and pretraining robots for performing different tasks \cite{Riedmiller2009}. In \cite{ernst2008reinforcement}, authors have shown on an electrical power oscillations damping problem that RL can be competitive with classic model-based methods, even when a good analytical model of the considered system is available. This lets to be optimistic about RL application for ancillary services. 

Ancillary services that involve a relatively small amount of energy to change power consumption rely on the capacity that is held in reserves, including thermostatically controlled loads \cite{DRAncillaryServices}, \cite{heffner2008loads}. The authors in \cite{claessens2018model} applied fitted Q-iteration RL method to obtain a performance within 65\% of a theoretical lower bound on the cost for a district heating network of 100 TCLs. This allows to be optimistic about RL application to optimisation of control that is applied to TCLs using other constraints, i.e. aiming demand-supply balance, not the cost optimisation.

To allow learning of the optimal control policy, interaction with a controlled system is required to gather experience for RL agent training. However, training an agent using a real power grid is not an acceptable option for an early stage of the development. Thus, the behaviour of a real power system is simulated with models developed in the Dymola environment using the Modelica language as an open access standard. The chosen power system model as a feeder of a number of TCLs represents the behaviour of interest that corresponds to properties of a real power grid. Specifically, being the most common type of load in the distribution grid thermostatically controlled loads (TCLs) can serve as means providing an ancillary service \cite{kirby1999load,heffner2008loads, meyn2015ancillary,zhang2012aggregate}. In \cite{tindemans2015decentralized} it was shown that an analytical approach to the controller development can be successful in demand response of thermostatic loads. One can achieve modulation of the power consumption of a heterogeneous set of TCLs according to a reference power profile.

In \cite{bogodorova2016voltage}, it was shown that voltage control-based ancillary services can be utilized using TCLs thermal capacity when regulating the power consumption using a voltage signal. However, the chosen constant control served as proof of concept. Therefore, more sophisticated control was mentioned as a future research direction. To expand this research, this paper focuses on the development of more complicated robust control. To find the required optimal voltage change-based control policy, reinforcement learning methods were applied.

The authors in \cite{moriyama2018reinforcement} achieved 22\% improvement in energy consumption compared to a model-based control of the data centre cooling model using deep RL techniques. In \cite{smottahedi-rl} the researcher applied Deep Reinforcement Learning to learn optimal energy control for a building equipped with battery storage and photovoltaics. In these cases, a reinforcement learning agent was trained and tested using power system model simulation developed using Modelica tools. Possibility of coupling reinforcement learning and Modelica models was validated in \cite{lukianykhin2019modelicagym}, where the authors built a pipeline for RL agents that are training in environments simulated with Modelica-compiled FMUs.

In \cite{ruelens2016reinforcement}, the authors succeeded to reduce the total cost of energy consumption of the single electric water heater by 15\% in a 40-days experiment. In addition, the authors emphasized the importance of proper state space discretization and dimensionality reduction, while using autoencoder for this purpose.
Moreover, in the review \cite{vazquez2019reinforcement} of applying mostly single agent methods and simplest algorithms such as Q-learning for demand response, the authors detected a tendency for better results, when action-state discretization and dimensionality reduction of action-state representations are applied. Although the authors investigated around 150 works, many of which applied control to TCLs, the control goal was mainly in cost reduction.

This paper describes an application of reinforcement learning to develop a voltage controller to manage power consumption of thermostatically controlled loads as part of ancillary services. The power system is modelled using Modelica that was integrated within ModelicaGym toolbox \cite{lukianykhin2019modelicagym} with Q-learning algorithm to train the controller.
This integration was done using Python and allows utilization of SOTA RL algorithms via OpenAI Gym.
The following project achievements are presented in the paper as contributions:
\begin{itemize}
    \item An optimal control for the considered voltage controller design (see Section~\ref{sec:app-model}) was learnt using Q-learning algorithm. The algorithm's hyperparameters tuning was done to improve its performance. The achieved performance is comparable to optimal constant control chosen using a simulation of a whole time interval.
    \item The developed controller shows the capability to generalize: to perform well on testing intervals longer than training.
    \item The experiment setup configurations were tuned to investigate particularities of the developed controller. These include changes of length and start time of a simulation used for training and testing a controller; setting of reference power level profiles.
    \item Dependency of controller performance on different environment state discretization strategies was investigated. These smart discretization strategies account for problem formulation and gathered historical data. The optimal width of bin detection method was tested. The dependency of the controller performance on the number of bins in equal-width binning was investigated.
\end{itemize}

The remainder of this paper is structured as follows. Section \ref{sec:prob} formulates the research problem and describes a Modelica model for which the problem is formulated. Section \ref{sec:approach} describes the approach to the controller development that was employed in experiments, including specific details of the utilized RL algorithm. It is followed by Section \ref{sec:exp} that describes conducted experiments, results that are supplemented with discussion.
Paper is finished by a short summary of the presented results together with conclusions and discussion of possible directions for future work in Section \ref{sec:con}.

\section{Problem Formulation}\label{sec:prob}

\begin{figure}
\includegraphics[width=0.5\textwidth]{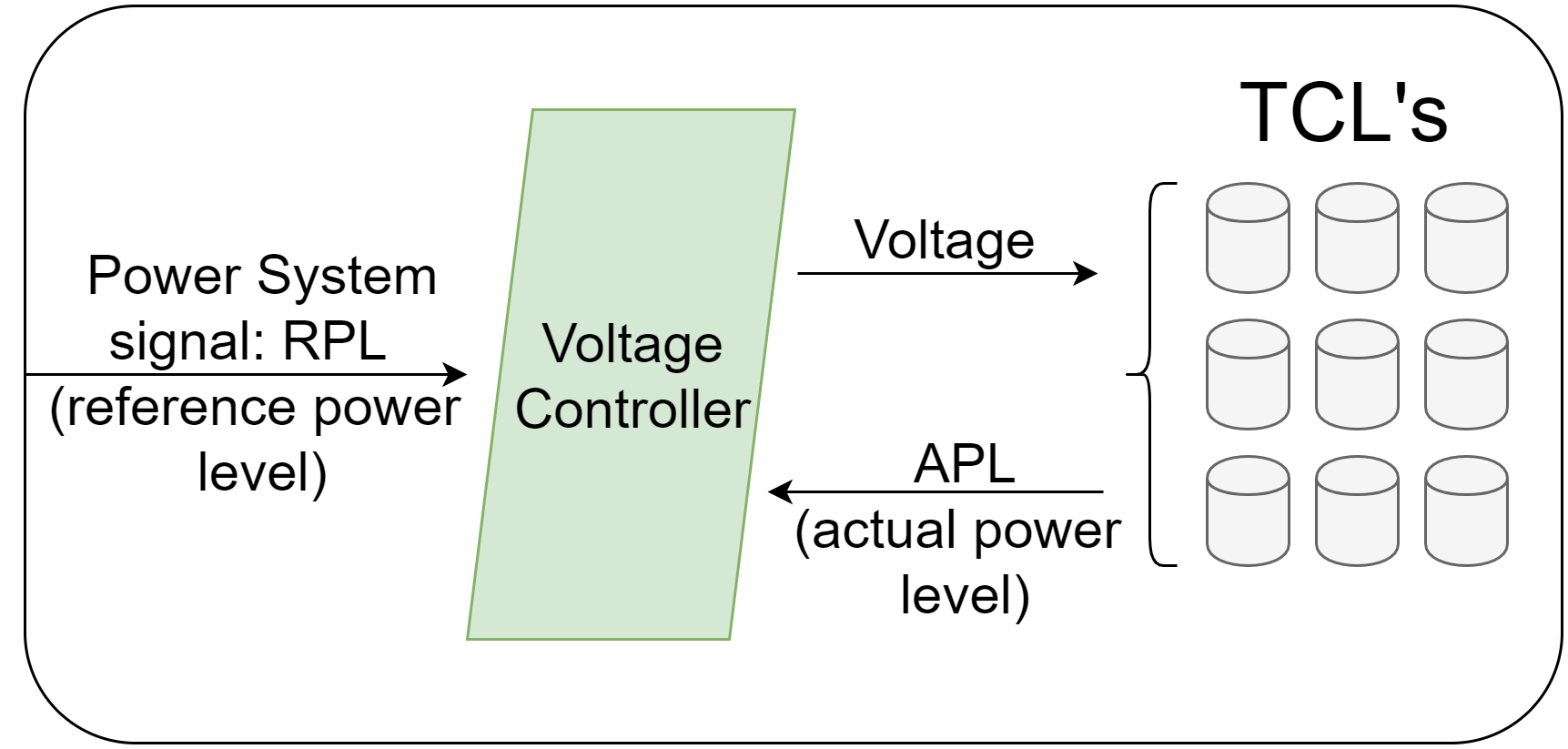}
\caption{Schematic picture of the considered power system configuration and controller to be developed} 
\label{fig:goal}
\end{figure}

More specifically, the considered applied problem can be formulated as follows:
given a grid model with multiple TCLs and means for voltage control, an optimal control strategy that controls power consumption at the point of common coupling should be found using RL (see Figure~\ref{fig:goal}).

The controller is placed in the point of common coupling of TCLs and controls the voltage at the substation. This way, sufficient load change can be achieved by varying voltage in a small range when satisfying power grid constraints \cite{bogodorova2016voltage}. Detailed controller design description can be found in \cite{bogodorova2016voltage}. 

The control goal is to make actual power consumption close to the reference power profile by changing the voltage on the bus. 
The voltage controller has one control parameter - proportional coefficient $k$.
A reinforcement learning agent should choose its value to change the voltage, taking into account current values of actual and reference power levels.
The control action is changed at discrete time steps of the considered time interval.

Mean squared error between actual and reference power levels is chosen as a measure of control strategy performance. This choice aimed to encourage control policies that avoid big differences between actual and reference power levels (APL and RPL, respectively). APL and RPL are sampled with the same time steps as control action is changed.

\subsection{Model}
\label{sec:app-model}
\begin{figure}[htb]
\centering
\includegraphics[width=0.5\textwidth]{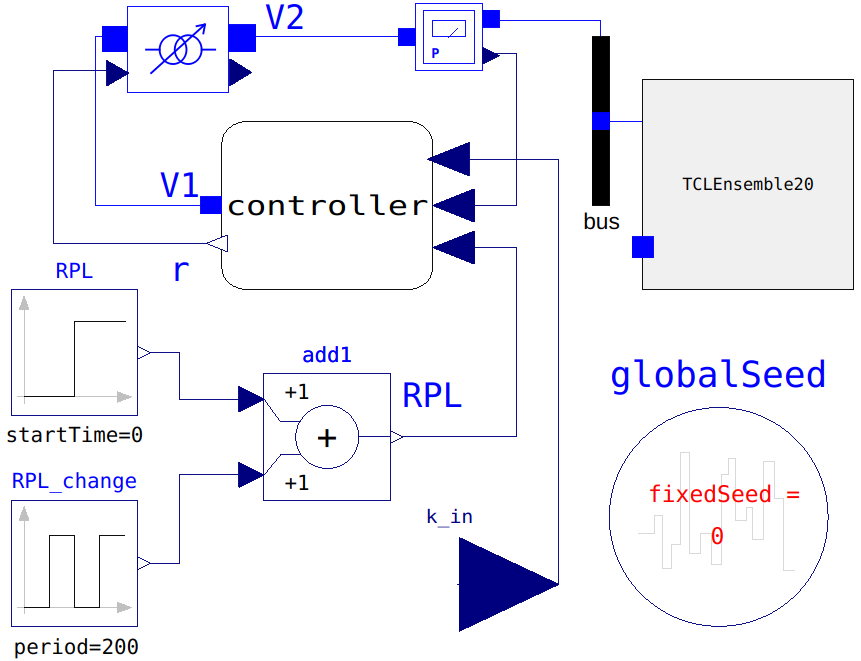}
\caption{OpenModelica diagram of a simulated power system (20 TCLs)}
\label{fig:sys-modelica}
\end{figure}

The considered power system model setup includes a feeder of 20 TCLs, tap changer, proportional controller that were modeled using Modelica language and Dymola environment (see Figure~\ref{fig:sys-modelica}). FMUs were compiled in JModelica, while OpenModelica was used for model diagram visualization.

Each TCL has the same thermal resistance $R=200^{\circ}$C/kW, power consumption $P=0.14$ p.u, ambient temperature $\theta_{a}=32^{\circ}$C, switching temperature range of $[19.75..20.25]^{\circ}$C and two variables: $switch$ indicating if TCL is on ($0$) or off ($1$) and temperature $\theta$. TCLs ($n=20$) differ in value of thermal capacitance $C$. Value of $C$ for $i$-th TCL is the $i$-th value in the array $[2.0$, $2.2286$, $2.4571$, $2.6857$, $2.9143$, $3.1429$, $3.3714$, $3.6$, $3.8286$, $4.0571$, $4.2857$, $4.5143$, $4.7429$, $4.9714$, $5.2$, $5.4286$, $5.6571$, $5.8857$, $6.1143$, $6.3429]$. 
Since thermostats in the grid are heterogeneous in their characteristics and start their operation at random moment of time, they are modeled to be initialized both deterministically (Equation~\ref{eq:theta-det}) or stochastically (Equation~\ref{eq:theta-stoch}). 

For the deterministic case differential equation of an individual TCL:
\begin{equation}
\label{eq:theta-det}
    \frac{d\theta}{dt}=\frac{-\theta_a+\theta+R\cdot P}{R \cdot C},    
\end{equation}

For the stochastic case the stochastic component $u$ is added as an input:
\begin{equation}
\label{eq:theta-stoch}
    \frac{d\theta}{dt}=\frac{-\theta_a+\theta+R\cdot P}{R\cdot C+R\cdot u \cdot range},
\end{equation}
where $u$ is a random number in $[0;1]$, $range=4.5$ - a range of thermal capacitance of $[C, C+range]$.

At the beginning of the simulation TCLs 1-10 are on ($switch=1$), TCLs 11-20 are off ($switch=0$). Equation~\ref{eq:switch} describes $switch$ value change. When TCL is on and temperature $\theta>\theta_{max}$ it switches off and when TCL is off and $\theta<\theta_{min}$ it switches on.
\begin{equation}
\label{eq:switch}
switch = 
\begin{cases}
    1, \quad \theta < \theta_{min},\\
    0, \quad \theta > \theta_{max}
\end{cases}
\end{equation}
Power consumption by each TCL:
\begin{equation}
P=switch \cdot g_0 \cdot v^2
\end{equation}
where $v$ - voltage, $g_0$ - conductance
\begin{figure}[htb]
\centering
\includegraphics[width=0.5\textwidth]{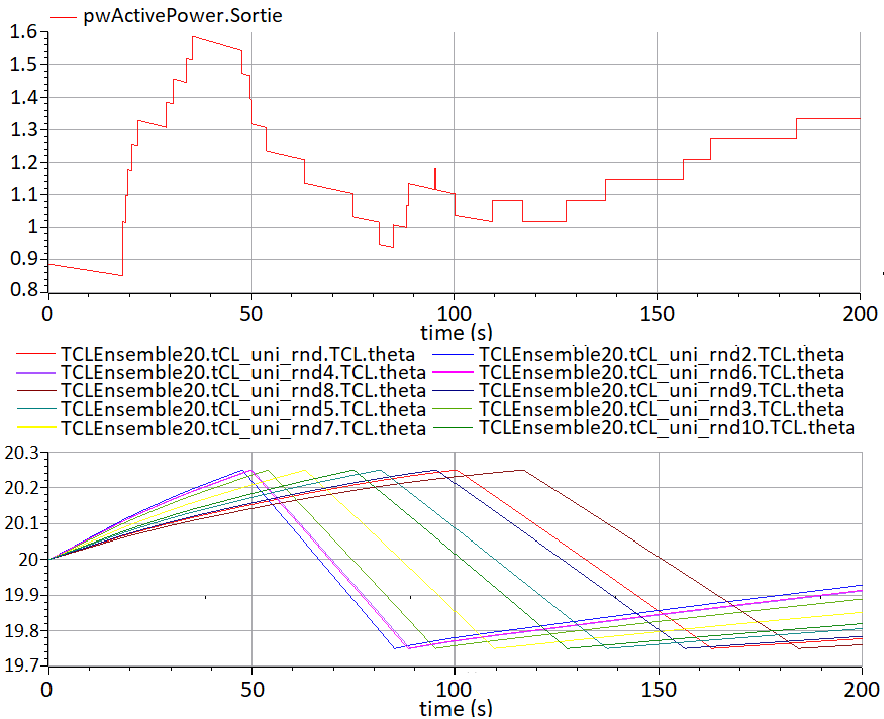}
\caption{Output of a simulated power system (20 TCLs)}
\label{fig:TCL}
\end{figure}

\section{Approach to Solution}
\label{sec:approach}

To investigate and assess opportunities for a voltage controller development using reinforcement learning, case studies should differ by their complexity. Therefore, to study RL controller's behaviour, deterministic TCL parameters initialization case and stochastic initialization were considered.  

To verify the scalability and ability to generalize of the controller, several case studies were carried out: 
\begin{itemize}
    \item step down in reference power level (RPL) in the middle of the considered time interval;
    \item simulation interval for testing is longer than the one used for training;
    \item beginning of training of the controller at different start time for a simulation: including or excluding transition process at the beginning of a simulation.
\end{itemize}

Since the transition process is prominent for the particular model (Figure \ref{fig:TCL}), it is valuable to investigate the influence of a transition process on a controller training and its performance. A step down in RPL in the middle of the considered time interval ($t=200s$) allows to verify an ability of the RL controller to follow the changing power consumption reference. 

Most experiments focus on the case with constant RPL and stochastic TCLs' parameters initialization. This set up allows to validate the RL controller at the early development stage and has a possibility to scale to other model configurations, e.g. changes in RPL.

Q-learning algorithm with straightforward rewarding strategy was chosen to train an RL controller. In addition, binning was utilized as a basic discretization technique to handle continuous state and action spaces. 




\subsection{Q-learning for Modelica model}

The Q-learning procedure that is utilized for RL controller training was presented in \cite{lukianykhin2019modelicagym}. The procedure is a training of a Q-learning RL agent in an environment simulated with a Modelica model. It is implemented in Python and makes use of OpenAI Gym via ModelicaGym toolbox.
It includes executing an action in the environment and processing corresponding feedback to learn optimal control with a Q-learning algorithm. The following hyperparameters have to be tuned for Q-learning algorithm: number of episodes of training, learning rate, exploration rate, exploration rate decay, discount factor, available actions and rewarding strategy.
Size of a time step between two consecutive control signal changes is a hyperparameter of the experiment as well. The time step size is set when a simulation of the environment is initialized for the experimental procedure. Size and number of such steps define the length of the time interval simulated in the experiment for controller training and testing. Unless otherwise specified, the considered time interval length is equal to 200 seconds and time step is equal 1 second, the chosen number of training episodes is 100.

In addition, for all experiments, controller training was repeated to gather statistical data for more reliable estimates of a controller's performance and evaluation of corresponding system behaviour. Unless otherwise specified, the number of the repeated experiments equals 5. For the training, a trend in the controller's performance was estimated. To this end, the episode performances were smoothed with a window of size 20 for each repeat and averaged over repeats. The received average smoothed performance that depends on a training episode is a proxy for performance change evaluation during training.
For the testing part of each repeat, the performance of the trained controller was estimated by repeated exploitation of a learnt RL-driven control strategy. Unless otherwise specified, the number of test episodes where performance is measured equals 50.

\subsection{Binning of a continuous state}
\label{sec:app-discr}
Q-learning, like most other RL algorithms, considers discrete environment states and actions available to an agent. On the contrary, real-world applications, like the problem considered in this paper, deal with continuous domains. Thus, there is a strong need in converting continuous spaces to reasonable discrete representation. The approach used for this purpose is called a discretization strategy. 

As the considered problem is not high-dimensional, it was decided to focus on the application of a classic discretization strategy - binning. The main idea of binning is to divide continuous space into numbered intervals (bins) and encode data by an index of the bin corresponding to the data point. Different strategies for choosing bins edges can be utilized in binning.

Several strategies for choosing discretization bins were tested:
\begin{enumerate}
    \item classic equal-width interval splitting;
    \item optimal width estimation methods (analogy with histograms);
    \item historical data quantiles as bin edges;
    \item accounting for the reference power level, when choosing bin edges.
\end{enumerate}

A basic approach to discretization of a continuous variable using bins is to split the space of possible values in bins of equal width. To account for values outside the interval, the most left bin can be half-opened from the left, the most right one can be half-opened from the right. The number of bins is a hyperparameter of this discretization strategy. Width of the bin is found by dividing the length of the interval of possible values by the number of bins. 

The second strategy is when an optimal number of bins is estimated. For this purpose, optimal width detection methods were used from the problem of building a histogram. For example, the Freedman-Diaconis optimal bin width estimator \cite{diaconis1981histogram} shows robustly reasonable results for large datasets. Besides, historical data is required to use this optimal-width estimation approach.

In third, it was decided to account for the RPL, when discretizing space for APL, by making RPL an edge of a bin. This way, possible RL agent confusion is avoided because all values of APL higher than the RPL are encoded differently from values that are lower. Other bin edges in this method were chosen using the equal-width approach.

Historical data allows to apply one more method for choosing bins: use data quantiles as bin edges. In this case, the obtained bins are not of equal width. This way, less attention is paid to parts of the space with a low number of observations, while providing a detailed representation for regions with a high concentration of observations.


\section{Experiments}
\label{sec:exp}
\rowcolors{2}{gray!5}{gray!25}
\begin{table*}[htb]
    \caption{Performance of the baseline and the competing approach for deterministic TCL parameters initialization case (best results in bold)}\label{tab:bas-det}
    \centering
    \begin{tabular}{|c|c|c|c|c|} 
    \toprule
    & \multicolumn{2}{c}{\emph{Constant RPL(1.2 p.u.)}} & \multicolumn{2}{|c|}{\emph{Step down in RPL(1.4-1.1 p.u.)}}\\
    \midrule
    \emph{\shortstack{Value of\\ a control parameter k}} 
    & \shortstack{Time step 1s,\\ \textit{MSE}}  & \shortstack{Time step 5s,\\ \textit{MSE}}
    & \shortstack{Time step 1s,\\ \textit{MSE}} & \shortstack{Time step 5s,\\ \textit{MSE}}\\
    \midrule
    Baseline & 0.1465 & 0.1382 & 0.2055 & 0.1983 \\
    0.5 & \textbf{0.0806} & \textbf{0.0767} & \textbf{0.1145} & \textbf{0.1138}\\
    1 & \textbf{0.0806} & \textbf{0.0767} & 0.1147 & 0.116\\
    2 & 0.0909 & 0.0896 & 0.1407 & 0.1494\\
    3 & 0.0917 & 0.0789 & 0.1395 & 0.1299\\
    4 & 0.0910 & 0.0962 & 0.1394 & 0.1531\\
    5 & 0.09 & 0.0834 & 0.1405 & 0.1298\\
    6 & 0.0913 & 0.1 & 0.1350 & 0.1418\\
    7 & 0.0894 & 0.1 & 0.1334 & 0.1402\\
    \bottomrule
    \end{tabular}
\end{table*}
First, case studies are described and their performance is evaluated. For the deterministic initialization case, both constant and step down in RPL, time steps of $t=1s$ and $t=5s$ were considered. For the stochastic initialization results for constant RPL and time step of $t=1s$ are presented.

Second, Q-learning was applied to find an optimal control policy. For the deterministic case with constant RPL and time steps of $t=1s$ and $t=5s$, hyperparameters are tuned, and the influence of parameters change are studied. For the stochastic case with constant RPL, time steps $t=1s$ and $t=5s$, previously chosen hyperparameters were utilized. Afterwards, tuning of the hyperparameters for time step $t=1s$ was continued. The dependency of the controller's performance on RPL and simulation interval length was studied. In addition, step down in RPL for $t=1s$ setup was used to validate the ability of the controller to generalize by testing it on longer time intervals than training.

Third, case study of different continuous state space discretization strategies is presented, including investigation of dependency of performance on the number of bins in equal-width binning and application of smart discretization strategies described in Section~\ref{sec:app-discr}. In this case, only constant RPL and time step $t=1s$ were considered.

\subsection{Alternative approaches}

Alternative approaches were determined to have a reference for the developed RL controller performance evaluation. Their performance was measured and used to evaluate the trained controller. These approaches are no controller in the system (baseline) and optimal constant control applied using the same controller design as for RL-learnt control (here and further called a competing approach).

\rowcolors{2}{gray!5}{gray!25}
\begin{table}[htb]
    \caption{Performance summary (median, mean, std) for the baseline (no control) and the competing approach (optimal constant control) for stochastic TCLs parameters initialization, $1s$ control change, constant RPL of 1.2 (best results in bold)}
    \label{tab:bas-stoch}
    \centering
    \begin{tabular}{|c|c|c|c|} 
    \toprule
    \emph{\shortstack{Control param. k}} & Median & Mean & Std\\
    \midrule
    Baseline & 0.0546 & 0.0613 & 0.0196 \\
    0.5 & 0.0396 & 0.0421 & 0.0085 \\
    1 & 0.0427 & 0.0424 & 0.0094 \\
    2 & 0.444 & 0.0447 & 0.0098 \\
    3 & 0.0366 & 0.0379 & 0.0077 \\
    4 & 0.0263 & 0.0269 & 0.0053 \\
    5 & 0.0198 & 0.0208 & 0.0041 \\
    6 & 0.0154 & 0.0159 & 0.004 \\
    7 & \textbf{0.0126} & \textbf{0.013} & \textbf{0.0043}\\
    \bottomrule
    \end{tabular}
\end{table}

Resulted performance for both RPL configurations and time steps of $t=1s$ and $t=5s$ are presented in Table~\ref{tab:bas-det}. Constant control with control parameter value $k=0.5$ leads to the best performance. However, this case represents only one sample from the possible system realization space. Thus, a more generalized stochastic case should be considered.
It was observed that higher values of the error metric for the non-constant RPL case, comparing to the constant one (see Table~\ref{tab:bas-det}). It can be explained with the fact that applying constant control action is not the best strategy when the RPL is not constant.

\rowcolors{2}{gray!5}{gray!25}
\begin{table}[htb]
    \caption{Optimal hyperparameters for the deterministic case}
    \label{tab:opt-det}
    \centering
    \begin{tabular}{|c|c|} 
    \toprule
    \emph{Parameter} & \shortstack{Chosen\\ value} \\
    \midrule
    Learning rate & 0.5\\ 
    Exploration rate & 0.5\\
    Exploration rate decay & 0.9\\
    Discount factor & 0.6\\
    Actions & [0.1, 0.5, 1, 2, 7]\\
    Reward & squared error scaled by -1000\\
    \bottomrule
    \end{tabular}
\end{table}
For the stochastic initialization system behaviour is not the same under the same control. Therefore,  trajectories were sampled and the corresponding performance metric was measured 50 times. Case of a constant RPL and time step of $t=1s$ is presented in Table~\ref{tab:bas-stoch}. However, for each different setup (RPL, time step or time interval) reference for the developed controller evaluation was determined in the same way as in the presented examples.


It is observed that best performing constant actions also correspond to the lowest variance of the measured performance. In addition, an optimal for this setup control parameter value $k=7$ is different from the one for deterministic initialization case. The MSE for step down in RPL is higher compareded to the constant RPL case.  

The received controller's performance is compared with the competing approach which is times better performing than the baseline. Distribution of performance is taken into account during each comparison, when setup with a stochastic initialization is considered.

\subsection{Q-learning}
\subsubsection{Deterministic Case}

As a first step, Q-learning application experiments were run on the deterministically initialized model. The optimal hyperparameters search was performed resulting in the parameters set that is used by default for all further experiments. The search was done by fixing all parameter values, but changing one or several parameters that are tuned. Optimal hyperparameters values for the deterministic case are given in Table~\ref{tab:opt-det}.

\begin{figure}[htb]
\centering
\includegraphics[width=0.5\textwidth]{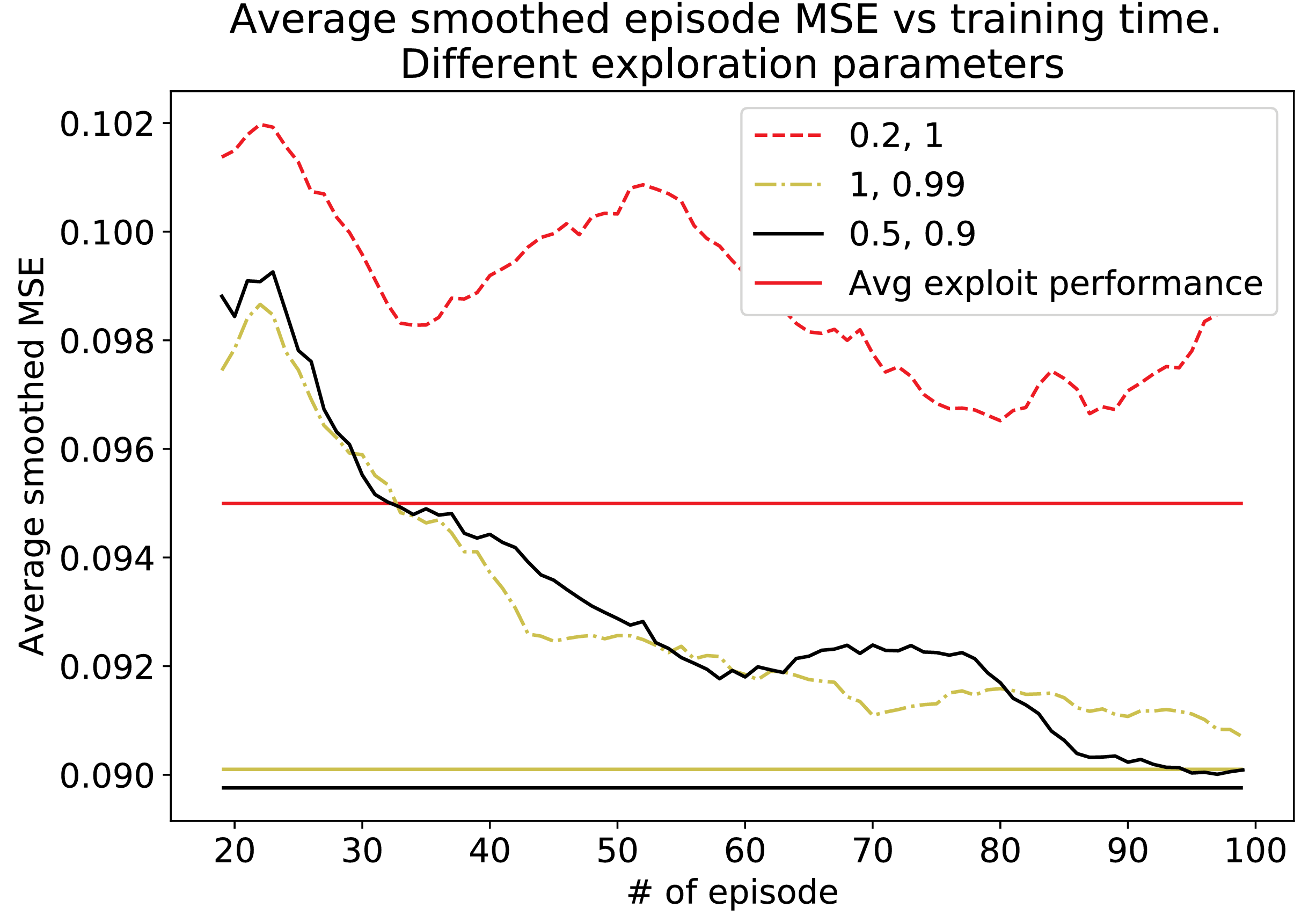}
\caption{Experiment to change of exploration parameters. Trend in performance during training (average smoothed MSE vs training episode)}
\label{fig:det-const-expl-exp}
\end{figure}
In some cases influence of a parameter on the performance was clear, while in exploration parameters variation experiment it wasn't obvious. The average smoothed episode performance is summarized in Figure~\ref{fig:det-const-expl-exp}. In Figure~\ref{fig:det-const-expl-exp} small exploration rate with big exploration rate decay is an obstacle for training to converge. At the same time, difference between two other parameters configuration is hard to observe, because training process looks almost the same. The average performances of the controller during testing are close to each other, difference in average MSE is less than $0.001$.  
\begin{figure}[htb]
\centering
\includegraphics[width=0.5\textwidth]{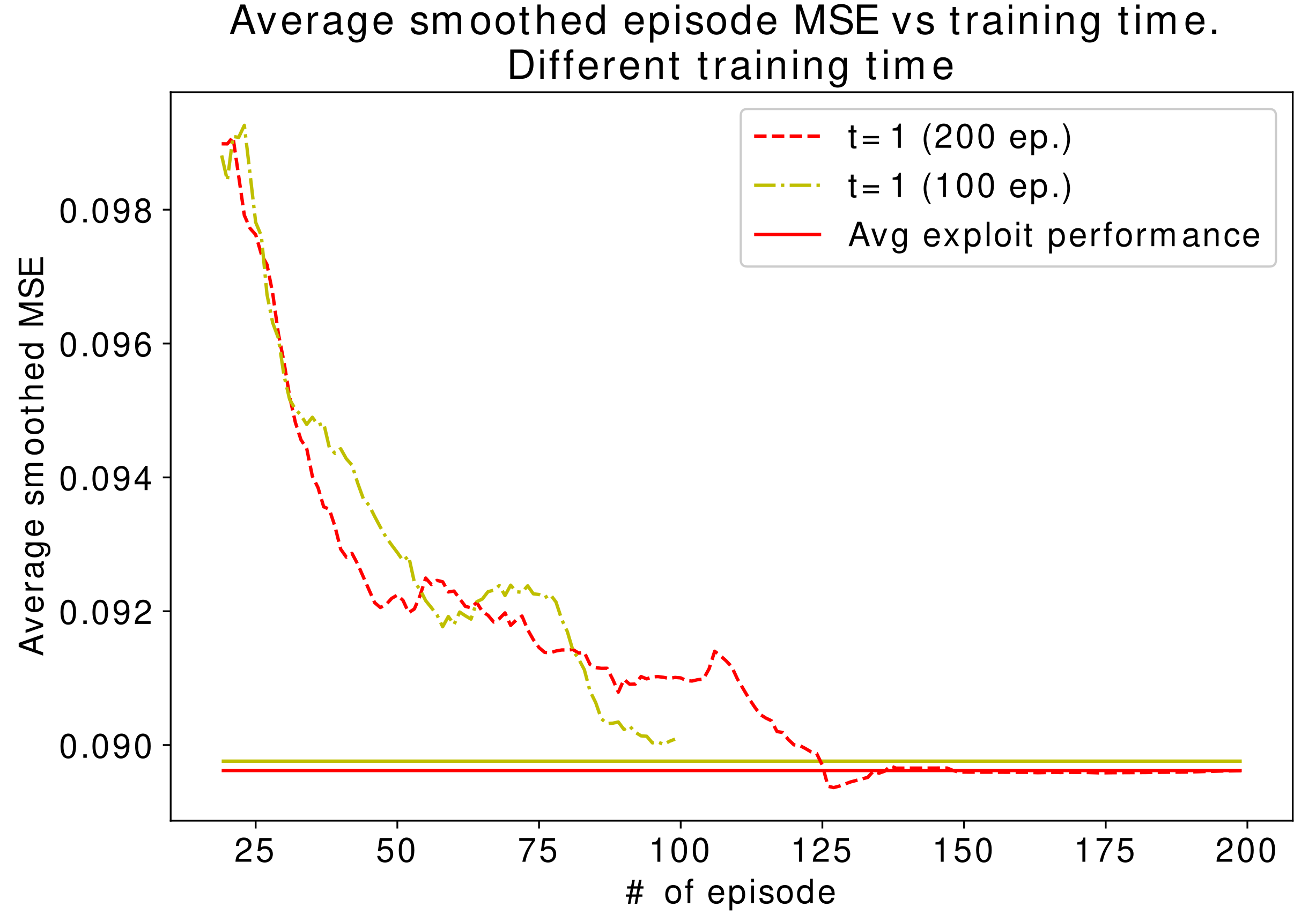}
\caption{Optimal hyperparameters, deterministic initialization experiment. Trend in performance during training (average smoothed MSE vs training episode)}
\label{fig:det-const-best-exp}
\end{figure}



After choice of the optimal hyperparameters, experiment was run for longer training time - 200 episodes (see Figure~\ref{fig:det-const-best-exp}). Longer training with the given parameters for time step of size $t=1s$ and $t=5s$ between control signal change is not leading to improvement in performance (see Figure~\ref{fig:det-const-best-exp}). The RL agent converges to a suboptimal strategy, because performance is lower than for a constant control action $k=0.5$. This can be caused by particularities of the system realization, as it is deterministicaly initialized being a single realization from the space of all possible system realizations. 
For generalization and effective agent's learning, experiments were continued on the stochastic case. 

\begin{figure}[htb]
\centering
\includegraphics[width=0.5\textwidth]{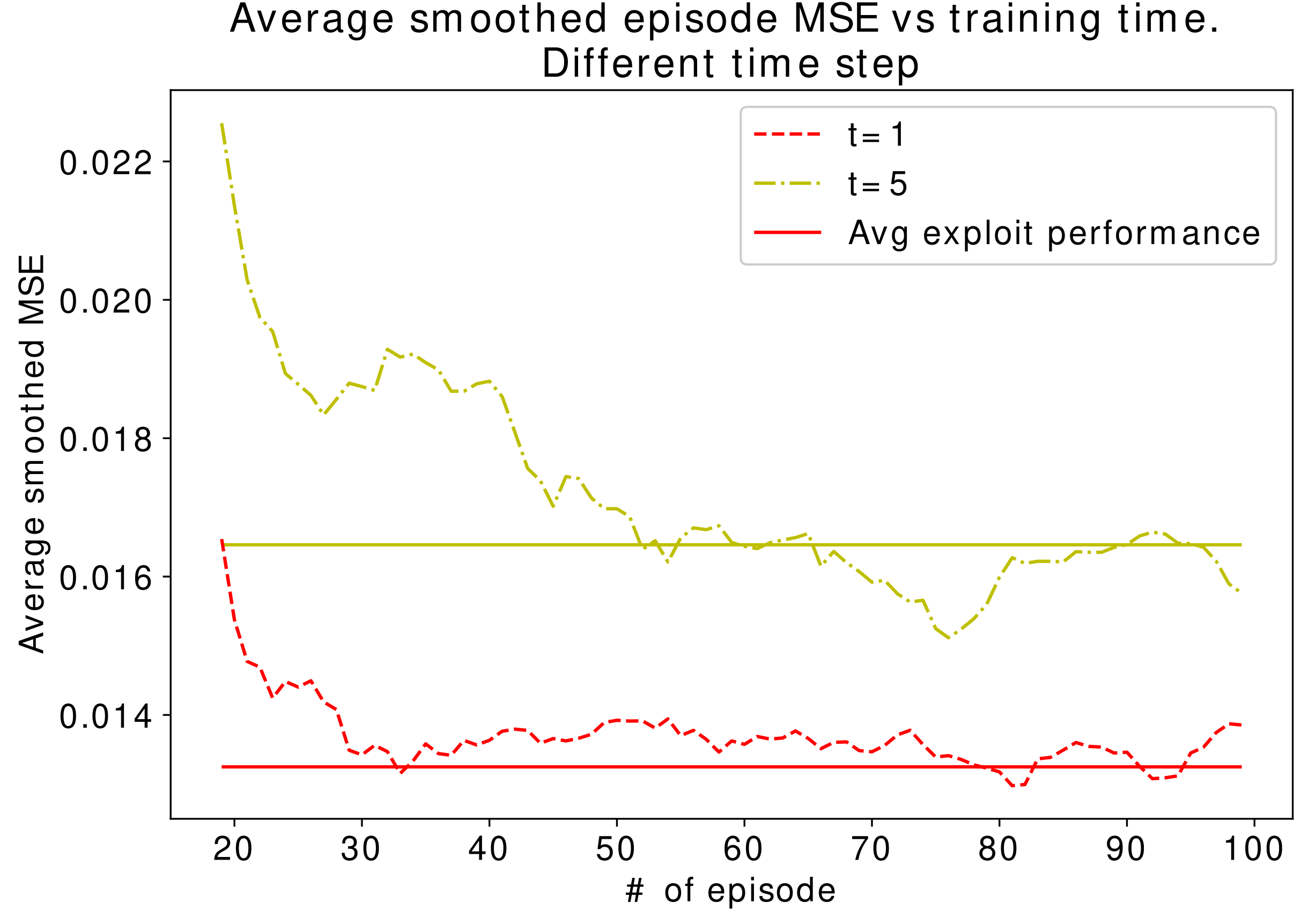}
\caption{Inferred optimal hyperparameters, stochastic initialization experiment. Trend in performance during training (average smoothed MSE vs training episode)}
\label{fig:stoch-const-best-exp}
\end{figure}

\subsubsection{Stochastic Case}
First, to introduce a stochastically initialized model, Q-learning experiments were performed for the constant reference power level. Second, step down in the reference power level was introduced at the half of considered time interval. For the both cases, experiments were run with the Q-learning utilizing optimal hyperparameters chosen on the deterministic case. These parameters served as an initial inference about optimal hyperparameters for the stochastic case. As a result, training of the controller converges (see Figure~\ref{fig:stoch-const-best-exp} for visualization for the constant RPL case).

\begin{figure*}[ht]
\centering
\includegraphics[width=0.95\textwidth]{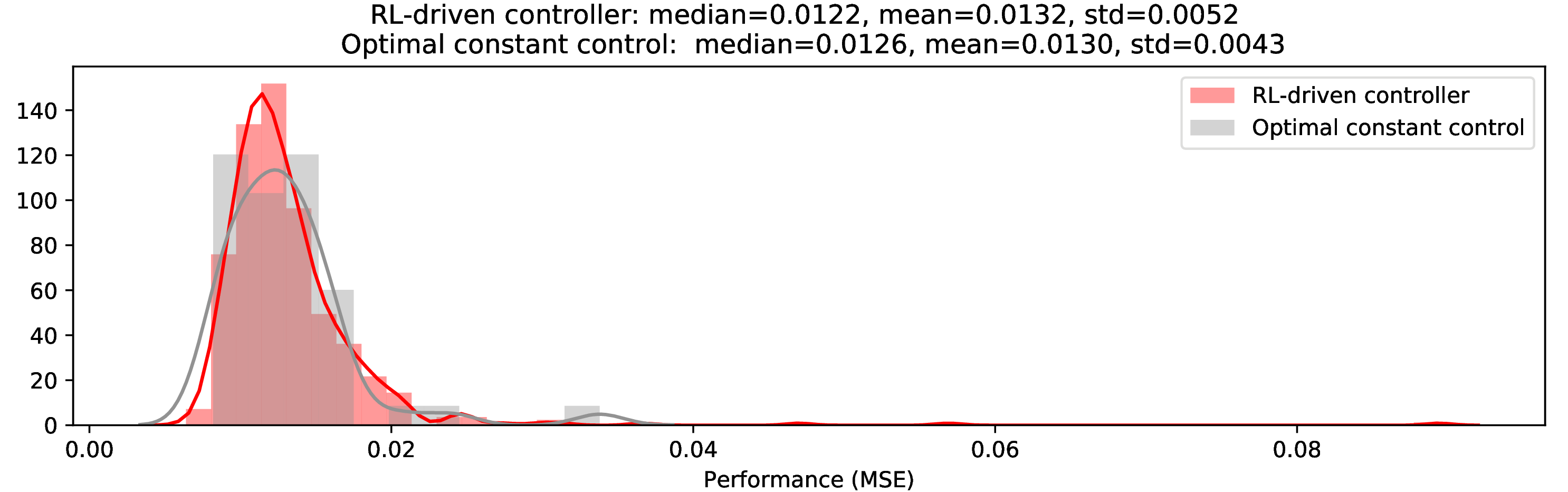}
\caption{Distribution of controller and the competing approach performance for constant RPL of 1.2 p.u.}
\label{fig:stoch-const-high-perf-dist}
\end{figure*}

\begin{figure}[ht]
\centering
\includegraphics[width=0.5\textwidth]{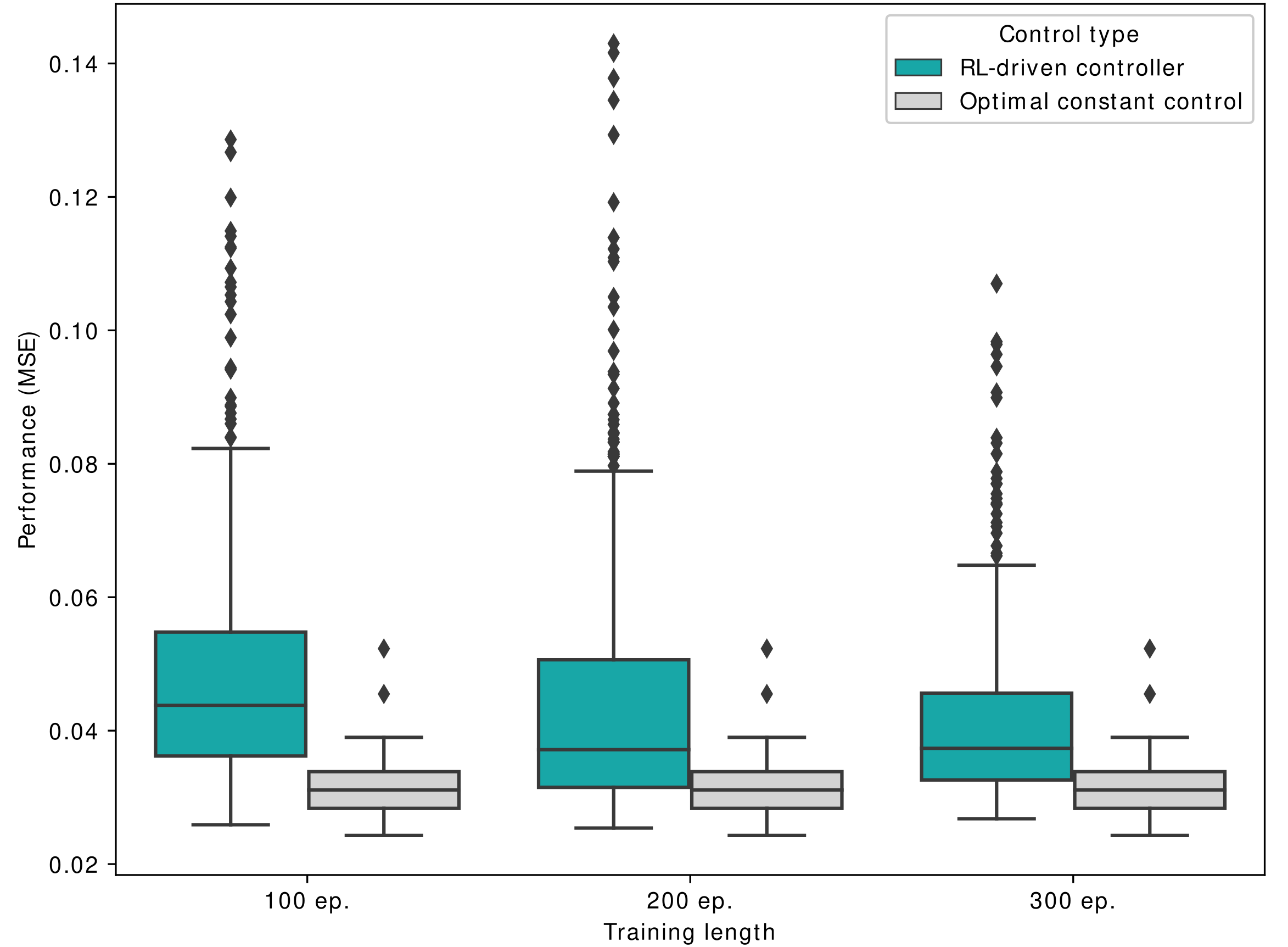}
\caption{Distribution of controller and the competing approach performance for different training time}
\label{fig:stoch-ampl-best-Q}
\end{figure}

\begin{figure*}[ht]
\centering
\includegraphics[width=\textwidth]{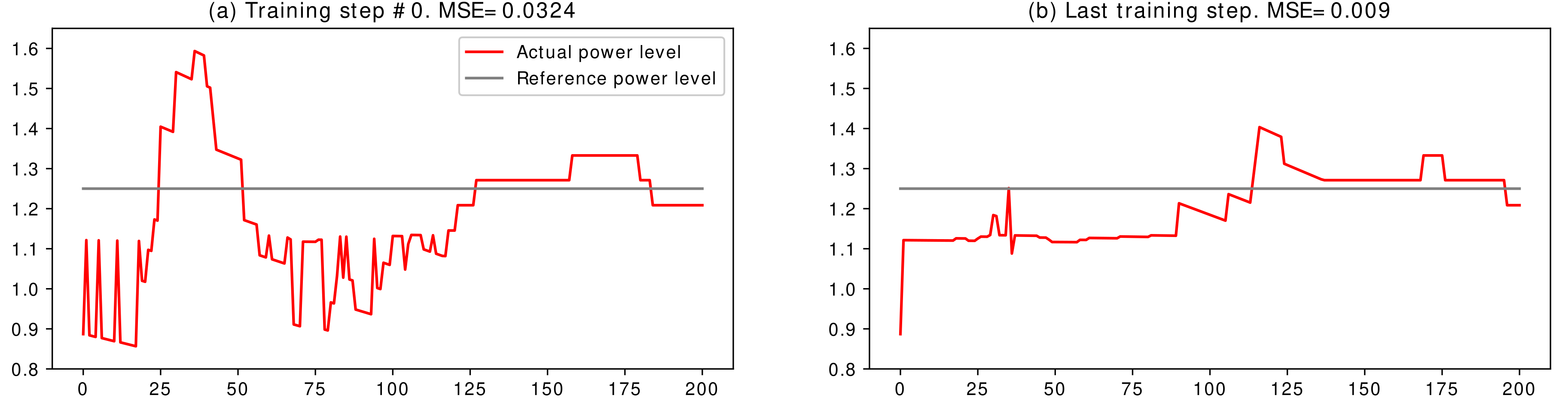}
\caption{System behaviour before (a) and after (b) controller training}
\label{fig:stoch-const-sys-beh}
\end{figure*}
\begin{figure*}[ht]
\centering
\includegraphics[width=0.95\textwidth]{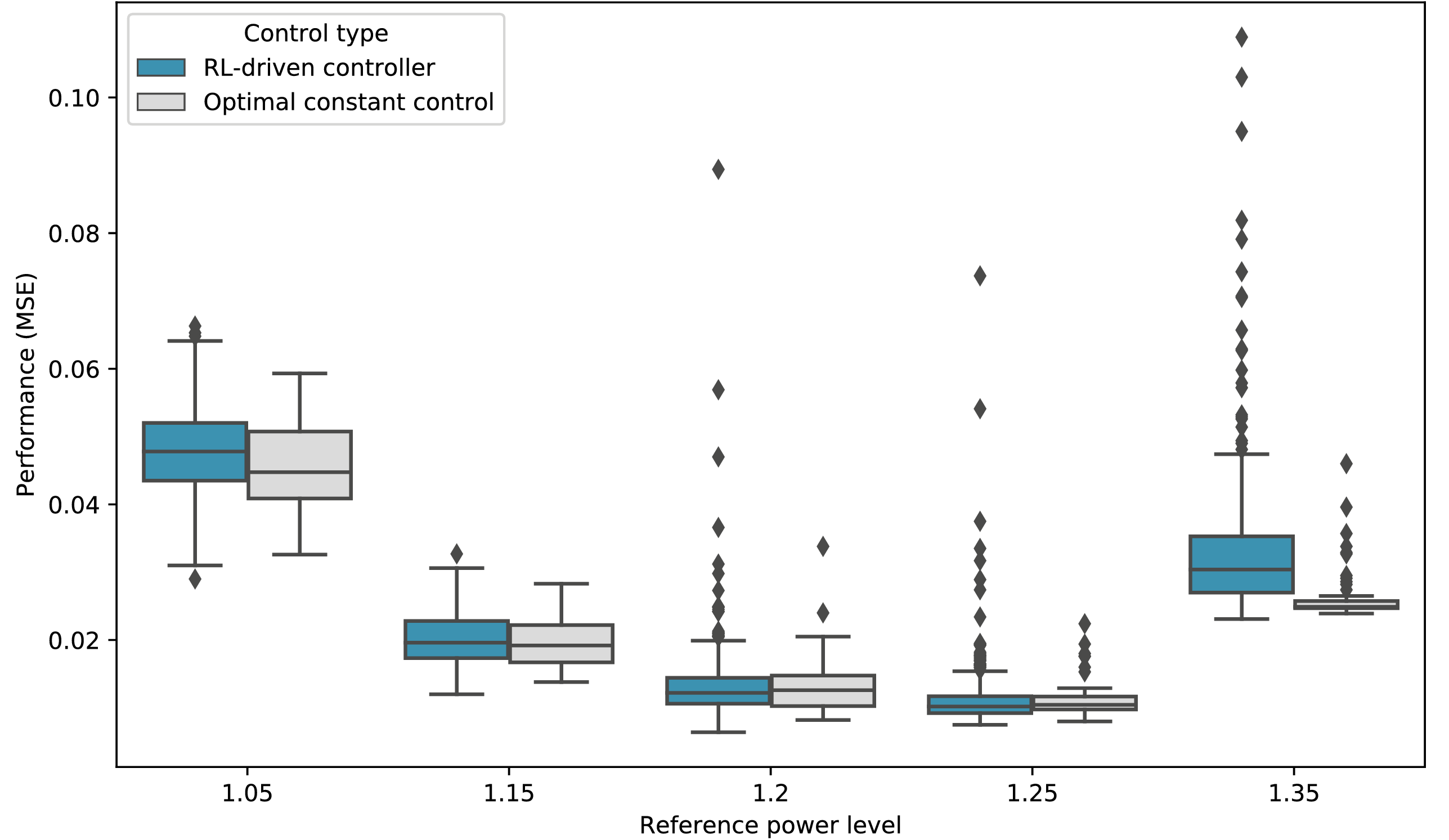}
\caption{Distribution of controller and the competing approach performance depending on constant RPL}
\label{fig:stoch-perf-vs-ref}
\end{figure*}
\begin{figure*}[ht]
\centering
\includegraphics[width=0.95\textwidth]{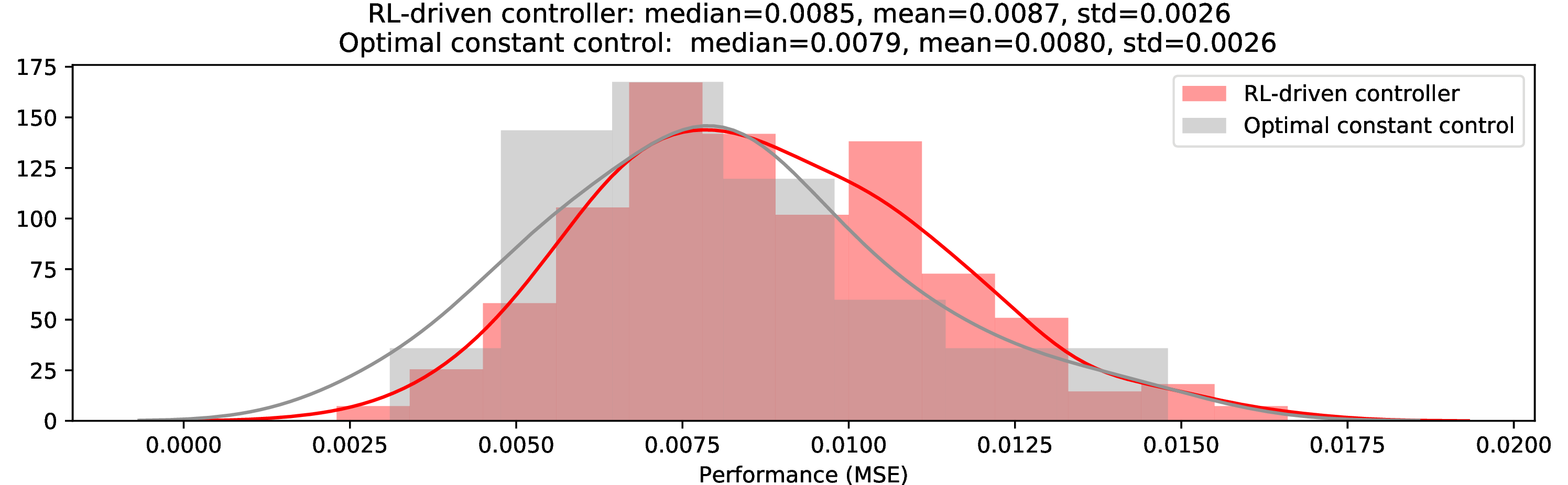}
\caption{Distribution of controller and the competing approach performance for constant RPL of 1.2 p.u. and time interval 175-375 seconds}
\label{fig:stoch-const-skip-175-perf-dist}
\end{figure*}
In Figure~\ref{fig:stoch-const-high-perf-dist} in the case with RPL of 1.2 p.u. RL-driven controller performs slightly better than the competing approach.
Figure ~\ref{fig:stoch-const-sys-beh} shows an example of system behaviour before and after controller training. It shown that APL approaches RPL after the RL-driven controller was trained, i.e. control goal is achieved at least to certain extent.

In Figure~\ref{fig:stoch-perf-vs-ref} to investigate the dependence of performance on the RPL, experiments were conducted for RPL in the range of $[1.05..1.35]\ p.u$. For all constant RPL values the trained controller demonstrates performance comparable to the competing approach, except of RPL of 1.35 p.u.

In addition, distribution of RL-based controller is skewed having a long right tail - presence of outliers with high MSE (see Figure~\ref{fig:stoch-const-high-perf-dist},~\ref{fig:stoch-perf-vs-ref}). To check if longer training can improve results, agent was trained on the same setup twice longer - for 200 episodes instead of 100. The longer training did not show capability to significantly improve exploitation performance of a trained controller, while being efficient for reducing variance of the controller's exploitation performance. It can be concluded that during the training RL agent reaches a plateau and converges to a (sub)optimal policy.

As hyperparameters were chosen based on the experiments for the deterministic case and received results are not beating the competing approach, an attempt to find better hyperparameters combinations for the stochastic case was made. However, no parameter changes led to significant improvements in the performance measures, while in some cases leading to decrease in the performance. Thus, this combination of hyperparameters is considered as optimal.

When the learnt policy is good in the long run, in the short run the better performance of the competing approach may be explained by particularities of the chosen performance metric and transition process that is observed in the system at the beginning of the considered time interval. As MSE puts stronger emphasize on big differences, one large-amplitude oscillation in APL at the beginning of a simulation may significantly influence results.
To test this hypothesis, experiment shown in Figure~\ref{fig:stoch-const-skip-175-perf-dist} was conducted on the time interval after transition process - $175-375$ seconds. It is observed, that there are no more outliers in controller performance. In this aspect, the performance of a trained RL-based controller is improved.

Q-learning controller training was done for the stochastic case with step down in RPL. In this case, to make a decision about a control action, controller takes into account both actual and reference power levels. The longer training did improved performance for control interval $t=1s$, but didn't improve for $t=5s$, so further investigation was done for the first option (see Figure~\ref{fig:stoch-ampl-best-Q}). It is observed that longer training may improve the average performance. However, at some point an increase in training time does not lead to improvements in average performance anymore, but reduces variance of it.

\rowcolors{2}{gray!5}{gray!25}
\begin{table*}[htb]
    \caption{Performance summary (median, mean, std) for Q-learning utilizing smart discretization strategies (HB - Histogram binning, CA - competing approach, HD - historical data, QL - Q-learning)}
    \label{tab:ext-discr}
    \centering
    \begin{tabular}{|c|c|c|c|c|} 
    \toprule
    n & Experiment name & Median & Mean & Std\\
    \midrule
    1 & 100 bins in $[0.1;1.7]$ & 0.0123 & 0.0132 & 0.0034\\
    2 & 10 bins in $[0.9;1.7]$ & 0.0126 & 0.0133 & 0.0036\\
    3 & 25 bins in $[0.9;1.7]$ & 0.0121 & 0.0131 & 0.0038\\
    4 & 50 bins in $[0.9;1.7]$ & 0.0126 & 0.0132 & 0.0038 \\
    5 & 100 bins in $[0.9;1.7]$ & 0.0127 & 0.0142 & 0.0066\\
    \midrule
    6 & \shortstack{HB CA HD} & 0.0123 & 0.0134 & 0.0045\\
    \midrule
    7 & \shortstack{HB QL HD (32 episodes)} & 0.0126 & 0.0136 & 0.0042\\
    \midrule
    8 & \shortstack{HB, CA HD (200 episodes training)} 
    & 0.0122 & 0.0131 & 0.0039\\
    \midrule
    9 & \shortstack{10 quantile bins, CA, HD} & 0.0128 & 0.0134 & 0.0039\\
    \midrule
    10 & \shortstack{10 quantile bins, QL, HD (32 episodes)} & 0.0126 & 0.0137 & 0.0049\\
    \midrule
    11 & \shortstack{10 bins, RPL as bin edge CA HD} 
    & 0.0123 & 0.0131 & 0.0037\\
    \midrule
    12 & \shortstack{10 bins, RPL as bin edge QL HD (32 episodes)} 
    & 0.0126 & 0.0132 & 0.0035\\
    \bottomrule
    \end{tabular}
\end{table*}

The results indicate that even simple RL methods are capable to learn control. The learnt control is comparable with the the competing approach utilizing the simulation of the whole considered time interval. At the same time, trained controller is not strongly superior to the competing approach in sense of performance, while can be considered superior from the point of view of generalization.

To test a generalization capability of the RL-driven controller, training on $t=200s$ interval with testing on $t=400s$ was considered. When the transition process $[0..175]s$ at the beginning of an experiment is skipped, it induces intervals of $[175..375]s$ and $[175..575]s$ for training and testing respectively. In this case, optimal constant control action that was chosen based on the training period is not the optimal one for the testing period. At the same time RL-driven controller's  policy allows the controller to slightly outperform the constant control

\subsection{Smart Discretization}

Dependency of the controller's performance on the number of bins in a fixed interval $[0.9;1.7]$ for APL was investigated and presented in Table~\ref{tab:ext-discr}. Training converged to almost the same performance in all cases. However, a bigger number of bins corresponds to a higher variance of the performance. This is due to higher state space dimensionality that slows down convergence of an RL agent.

To apply optimal bin width estimation approach, historical data was collected in the experiments for the competing approach and in early stages of Q-learning experiments with other discretization strategy. Both options for collecting historical data lead to almost the same performance that is shown in Table~\ref{tab:ext-discr}. The estimated number of bins is around 60. Because of that, longer training is required to achieve good performance and reducing its variance.

Afterwards, an approach using quantiles of historical data as bins edges was applied using the same historical data. For the considered $10\%$-quantiles results didn't improve in comparison to the previous discretization strategies (refer to Table~\ref{tab:ext-discr}). Most likely there is no improvement, because the bins are too big.

Next, accounting for RPL in APL binning, when using its value as a bin's edge, was utilized. Total number of bins equals 10 and different historical data were tested (see Table~\ref{tab:ext-discr}). The latter was used to determine approximate interval for discretization using maximum and minimum historical values. A combination with historical data from the competing approach showed the best results among all smart descretization techniques.

In addition, the experiments with different discretization strategies were conducted for the skipping transition case, i.e. $[175..375]s$ simulation interval. In this case all discretization strategies have a comparable performance.

Results of testing the discretization strategies lead to conclusion that it is advised to make smaller bins in more dense regions. Although good results can be achieved by exploiting a high number of bins in equal-width binning, a bigger number of bins slows down training convergence and leads to inefficiency, as some parts of a Q-table of such an agent are updated extremely rarely. 
Thus, the best approach is to account for problem formulation and use historical data to utilize smart discretization strategy.

\section{Conclusions \& Future Work}\label{sec:con}

The proof of concept developed in Python shows that RL is capable to learn efficient control policy to provide a voltage control-based ancillary service using TCLs. In addition, RL-based controller has shown the ability to learn efficient policies even without sophisticated tuning. Moreover, it shows the ability to generalize better than the competing approach of constant control. This way, a simulation of the whole considered interval can be avoided in practice. This is beneficial since such a simulation is not always available.

However, achieved performance, in general, is not significantly better than the performance of the competing approach. This may be explained by convergence to a suboptimal policy or impossibility to learn an optimal policy by considering only one time point in the system trajectory for making a control change decision. 
I.e. it could be due to impossibility for an RL-agent to capture all the dynamics in the system when only values of APL and RPL at one time point are considered.

As the main direction of future work, authors consider work on improvement of the controller performance by applying other RL algorithms, such as deep RL and batch RL techniques, Bayesian approaches in RL. 
In addition, it may be beneficial to consider problem formulation that includes more information than just APL and RPL at the considered time point for making a decision about control change. In addition, performance may be improved by introducing some changes in controller design, such as relaxing constraints for output voltage.

As a side note, it would be useful to parallelize experiments to enhance a further development process. An experiment with the RL-driven controller (includes training and validation repeated 5 times) takes about 6 hours, while it takes approximately 3.5 hours for the optimal constant control experiment (simulation of the considered time interval repeated for each of 8 possible control parameter values). Machine parameters were: Intel-i7 7500U 2.7GHz (3 cores available), 12GB RAM. At the moment, parallelization attempts have not achieved significant speed up. Although parallel experiments make use of additional available CPU cores and RAM, the simulation tools could be a bottleneck. 

\section*{Acknowledgements}
Authors would like to thank ELEKS (\url{https://eleks.com/}) for funding the Machine Learning Lab at Ukrainian Catholic University and this research.

\bibliography{main}


\end{document}